\documentclass[conference]{IEEEtran}

\usepackage{cite}
\usepackage[pdftex]{graphicx}

\usepackage{amsmath}
\usepackage{amssymb}
\usepackage{array}
\usepackage[caption=false,font=footnotesize]{subfig}
\begingroup\expandafter\expandafter\expandafter\endgroup
\expandafter\ifx\csname IncludeInRelease\endcsname\relax
  \usepackage{fixltx2e}
\fi
\usepackage{url}
\usepackage[bottom]{footmisc}
\usepackage{stfloats}
\usepackage{booktabs}
\usepackage{array}
\usepackage{setspace}
\usepackage{bm} 

\newcommand{\vect}[1]{\mathbf{#1}} 
\DeclareMathSymbol{\R}{\mathalpha}{AMSb}{"52} 

\begin{document}
%
\title{Recursive Multikernel Filters Exploiting \\ Nonlinear Temporal Structure}

\author{\IEEEauthorblockN{Steven Van Vaerenbergh}
\IEEEauthorblockA{Dept. Communications Engineering\\
University of Cantabria, Spain\\
steven.vanvaerenbergh@unican.es}
\and
\IEEEauthorblockN{Simone Scardapane}
\IEEEauthorblockA{DIET Dept.\\
Sapienza University, Italy\\
simone.scardapane@uniroma1.it}
\and
\IEEEauthorblockN{Ignacio Santamaria}
\IEEEauthorblockA{Dept. Communications Engineering\\
University of Cantabria, Spain\\
i.santamaria@unican.es}}


\maketitle

\begin{abstract}
In kernel methods, temporal information on the data is commonly included by using time-delayed embeddings as inputs. Recently, an alternative formulation was proposed by defining a $\gamma$-filter explicitly in a reproducing kernel Hilbert space, giving rise to a complex model where multiple kernels operate on different temporal combinations of the input signal. In the original formulation, the kernels are then simply combined to obtain a single kernel matrix (for instance by averaging), which provides computational benefits but discards important information on the temporal structure of the signal. Inspired by works on multiple kernel learning, we overcome this drawback by considering the different kernels separately. We propose an efficient strategy to adaptively combine and select these kernels during the training phase. The resulting batch and online algorithms automatically learn to process highly nonlinear temporal information extracted from the input signal, which is implicitly encoded in the kernel values. We evaluate our proposal on several artificial and real tasks, showing that it can outperform classical approaches both in batch and online settings.
\end{abstract}


\IEEEpeerreviewmaketitle

\section{Introduction}
\label{sec:intro}
In recent years, kernel adaptive filters (KAF) have become a popular approach for online machine learning and time-series prediction, thanks to numerous theoretical and practical advances \cite{vanvaerenbergh2012kernel,zhao2013fixed,yukawa2015adaptive}. Differently from deep recurrent neural networks \cite{goodfellow2016deep}, whose training is generally formulated in batch fashion, KAFs can be trained efficiently with a single pass over the training data. Popular examples of KAFs include kernel least-mean-square \cite{liu2008kernel,zhao2013fixed} and kernel recursive least-squares \cite{engel2004kernel,vanvaerenbergh2012kernel}. When operating on temporal data, most KAFs apply common kernel functions, e.g. Gaussian or polynomial, to time-delay embeddings of the input data. Choosing a specific embedding is not trivial in general, and it might be suboptimal in the case of non-stationary signals. These are known problems in other kernel methods as well, such as support vector machines (SVMs) \cite{mukherjee1997nonlinear} and kernel ridge regression (KRR).

In order to overcome these limitations, several authors have proposed kernelized extensions of classical recursive models, including the recurrent least-squares SVM \cite{suykens2000recurrent}, the autoregressive and moving average (ARMA) SVM \cite{martinez2006support}, the kernel machine and space projection (KMSP) method \cite{li2011identification}, the kernel $\gamma$-filter \cite{camps2004robust}, and, more recently, the kernel adaptive ARMA algorithm \cite{li2016kernel}. All of these works share a common methodology, which is composed of three major steps: (i) define a proper state-space model (SSM) in the input space; (ii) map the input and/or the state of the model to a high-dimensional feature map corresponding to a properly defined reproducing kernel Hilbert space (RKHS); and (iii) solve the resulting model by substituting all dot products with evaluations of the associated kernel function according to the so-called ``kernel trick''. Although this is a powerful methodology, it is hard to obtain insights from the model and, more importantly, selecting a proper embedding remains a crucial problem.

In this paper, we focus on the alternative methodology recently proposed in \cite{tuia2014explicit}. The basic idea consists in defining the SSM explicitly in a proper RKHS, where samples might not correspond one-to-one with the original inputs. In particular, it is possible to define a proper reproducing theorem such that the model can be expressed as a summation of kernel values, which can be computed \textit{recursively} at each time instant. For the specific case of the $\gamma$-filter, the resulting model is particularly appealing because each ``tap'' in the RKHS corresponds to a filtering operation on a different time-scale of the original input \cite{tuia2014explicit}. Unlike previously proposed recursive kernels, e.g. \cite{hermans2012recurrent}, this class of kernels was found to work robustly in many problems, including time-series prediction and  array processing. Some theoretical aspects of a related class of kernels were investigated independently in \cite{mouattamid2009recursive}.

Here, we are interested in extending this methodology by focussing on a shortcoming of the original model. In particular, standard kernel methods consider a single kernel value for each input datum, while in this case, we obtain several kernel values per input. In \cite{tuia2014explicit}, this problem was side-stepped by either averaging the values, or by computing inner products. This approach lacks in flexibility, however, because it implicitly assumes that all time-scales are equally important. In this paper, we put forth the idea of considering each kernel value as coming from a different kernel function, and to apply proper adaptive strategies to learn the dependency with respect to each of them. In the literature, the idea of adaptively combining kernel functions goes under the name of multiple kernel (MK) learning. MK algorithms originated in the SVM literature \cite{gonen2011multiple}, and their benefits have also been proven by a number of authors for KAFs, including the MK normalized LMS \cite{yukawa2012multikernel}, the mixture KMLS \cite{pokharel2013mixture}, the doubly regularized MKLMS \cite{yukawa2013online}, and Cartesian HYPASS \cite{yukawa2015adaptive}. 

In order to test the feasibility of our idea, we focus on a simple strategy in which multiple kernel estimators are linearly combined following a stacking-like \cite{breiman1996stacked} algorithm. This allows us to test the same formulation in both batch and online settings. In the experimental section, we show that by combining our procedure with the recursive kernel of \cite{tuia2014explicit}, we obtain a performance that is comparable to or better than competing approaches. To this end, we evaluate several benchmarks using both artificial and real-world datasets.

The rest of the paper is organized as follows. Section \ref{sec:recursive_filtering} describes the RKHS $\gamma$-filter from \cite{tuia2014explicit}, which is extended through the proposed MK strategy in Section \ref{sec:multikernel}. We briefly consider computational considerations of the recursive kernel evaluation in Section \ref{sec:efficient}. Experiments are presented in Section \ref{sec:experiments}, before giving some conclusive remarks in Section \ref{sec:conclusions}.

\section{Recursive $\gamma$-filtering in RKHS}
\label{sec:recursive_filtering}
\begin{figure}[t]
\centering
\includegraphics[width=.8\linewidth]{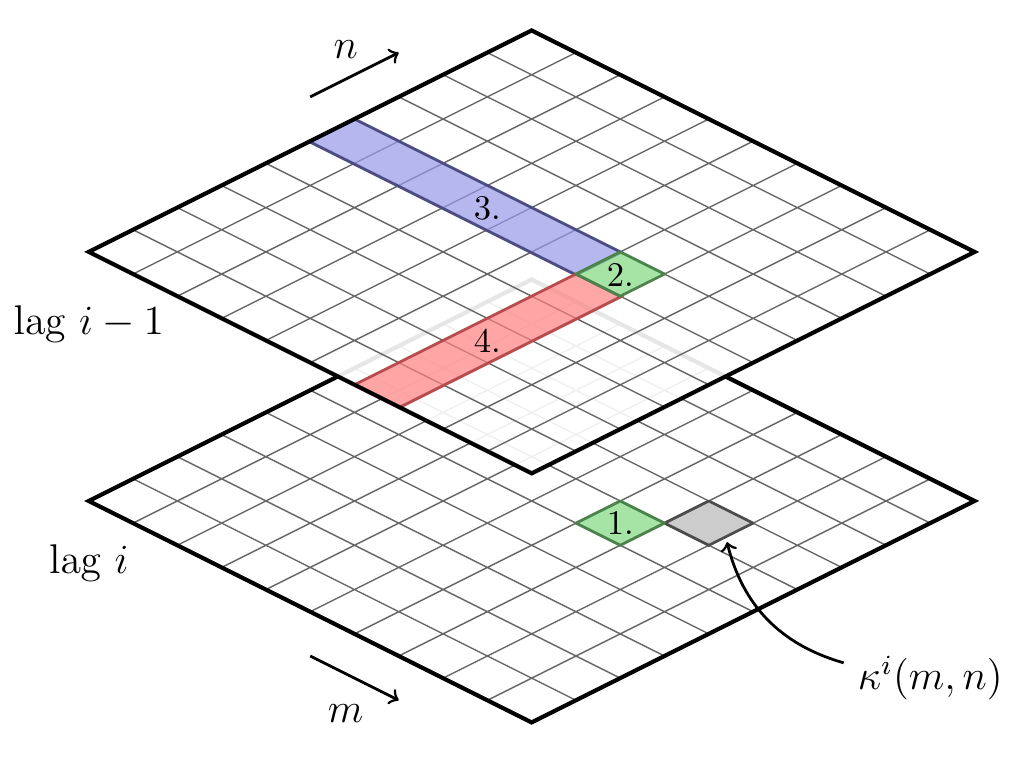}
\caption{Calculation of the recursive kernel matrix. The numbers $1$ to $4$ correspond to the four terms in Eq. \eqref{eq:recursive_kernel_definition}, $i>1$.}
\label{fig:recursive}
\end{figure}

Let us denote by $(x_n, y_n)$ a generic input-output pair observed at time instant $n$. We assume these data to be generated by the following nonlinear model
\begin{subequations}
\begin{align}
y_n & = f(\langle w^i, x_n^i \rangle) + e_y \,, \\
x_n^i & = g( x_{n-1}^i, x_{n-2}^i, \ldots, y_n, y_{n-1}, \ldots) + e_x \,,
\end{align}
\end{subequations}
where $x_n^i$ is the input signal at the $i$-th filter tap, $\langle \cdot \rangle$ denotes the inner product, $w^i$ are the filter weights, $f(\cdot), g(\cdot)$ are smooth nonlinear functions, and $e_x, e_y$ represent state and output noise respectively. The main idea introduced in \cite{tuia2014explicit} is to model a similar process, defined instead in a proper Hilbert space $\mathcal{H}$
\begin{subequations}
\begin{align}
y_n & = \hat{f}(\langle w^i, \phi_n^i \rangle_{\mathcal{H}}) + e_y \,, \\
\phi_n^i & = \hat{g}( \phi_{n-1}^i, \phi_{n-2}^i, \ldots, y_n, y_{n-1}, \ldots) + e_\phi \,,
\end{align}
\label{eq:general_model}%
\end{subequations}
where $w^i, \phi^i_n$ are now samples in the (possibly infinite-dimensional) $\mathcal{H}$, and $e_\phi$, $e_y$ represent the state and output noise. Differently from previous works, it is not required for $\phi_n^i$ to have $x_n^i$ as its preimage, in order to provide more flexibility to the model. Proving a representer theorem in general is not trivial, except for specific instantiations of Eq. \eqref{eq:general_model}. A particularly interesting case arises by assuming that the filtering operation in Eq. \eqref{eq:general_model} is a $\gamma$-filter \cite{principe1993gamma,camps2004robust} given by (omitting noise for simplicity)
\begin{subequations}
\begin{align}
y_n & = \sum_{i=1}^P \langle w^i, \phi_n^i \rangle_{\mathcal{H}} \,, \\
\phi_n^i & = \begin{cases}
	\psi(x_n) & \text{ if } i=1 \,, \\
	(1-\mu)\phi_{n-1}^i + \mu\phi_{n-1}^{i-1} & \text{ if } 2 \le i \le P
\end{cases}
\end{align}
\label{eq:gamma_filter}%
\end{subequations}
where $\psi(x_n)$ is some nonlinear transformation in $\mathcal{H}$, $P$ is the filter length controlling the memory depth, and $ 0 < \mu \le 1$ is a free parameter controlling stability. It is interesting to observe that each ``tap'' in the Hilbert space is defined by a recursive equation, so as to work over different temporal combinations of the (nonlinearly transformed) original input sequence. In \cite{tuia2014explicit}, it is proved that, for a given sequence of length $N$, the filter weights $w^i$ can be expressed as
\begin{equation}
w^i = \sum_{m=1}^N \beta_m^i \phi_m^i \,,
\label{eq:representer_theorem}
\end{equation}
for some coefficients $\beta_1^i, \ldots, \beta^i_N \in \mathbb{R}$. After substituting \eqref{eq:representer_theorem} in \eqref{eq:gamma_filter}, and making use of the kernel definition $\kappa^i(m, n) = \langle \phi_m^i, \phi_n^i \rangle_{\mathcal{H}}$,  we obtain
\begin{equation}
\hat{y}_n = \sum_{i=1}^P \sum_{m=1}^N \beta_m^i \langle \phi_m^i, \phi_n^i \rangle = \sum_{i=1}^P \sum_{m=1}^N \beta_m^i \kappa^i(m, n) \,.
\label{eq:y_hat_n}
\end{equation}
Again, it is worth underlining that, in general, $\kappa^i(m,n) \neq \kappa^i(x_m^i, x_n^i)$. In particular, due to the recursive model definition, the kernel itself can be defined recursively, and the closed-form expression is given by\footnote{Note that the original formula in \cite[Eq. (13)]{tuia2014explicit} was missing a summation from $j=2$ to $n-1$. The correct formula is given by Eq. \eqref{eq:recursive_kernel_definition}.}
\begin{equation}
\kappa^i(m,n) = \begin{cases}
	\kappa(x_m, x_n), & i = 1 \\
	\bar{\mu}^2 \kappa^i(m-1, n-1)\\
	~~ + \, \mu^2 \kappa^{i-1}(m-1, n-1)\\
	~~ + \, \mu^2 \displaystyle\sum_{j=2}^{m-1} \bar{\mu}^{j-1} \kappa^{i-1}(m-j, n-1)\\
	~~ + \, \mu^2 \displaystyle\sum_{j=2}^{n-1} \bar{\mu}^{j-1} \kappa^{i-1}(m-1, n-j), & i > 1
\end{cases}
\label{eq:recursive_kernel_definition}
\end{equation}
where $\bar{\mu} = 1 - \mu$, and $\kappa(x_m, x_n)$ is the classical (scalar) kernel function corresponding to $\langle \psi(x_m), \psi(x_n) \rangle_{\mathcal{H}}$. The calculation of the associated kernel matrix is illustrated in Fig. \ref{fig:recursive}. More generally, we can replace $x_m$ and $x_n$ with the corresponding time-delayed embeddings $\bm{x}_m$ and $\bm{x}_n$ to obtain a more expressive model with similar computational complexity. 

The recursive kernel $\gamma$-filter generalizes several known models, such as the classical $\gamma$-filter, recursive auto-regressive filters, and several others (see \cite[Section III-C]{tuia2014explicit}). Differently from standard KRR and KAF algorithms, which require one kernel value for each time instant, the model structure in \eqref{eq:y_hat_n} requires $P$ kernel values, each of which can be seen as operating at a different time-scale. The filtering approach proposed in \cite{tuia2014explicit} did not adapt all coefficients corresponding to all kernel values, but employed a simple combination of the kernels instead. One such combination is the composite kernel $\kappa(m,n) = \frac{1}{P}\sum_{i=1}^P \kappa^i(m,n)$. However, this approach has a drawback, namely, it assumes that all kernels are equally significant, and it may lose important information when combining them. In the next section, we will describe a more principled approach to combine the different kernels.

\section{Proposed adaptive multi-kernel approach}
\label{sec:multikernel}
We now describe an adaptive formulation that automatically weights the different kernels. The proposed algorithm is relatively inexpensive, and it can be implemented easily in both batch and online settings. Note, however, that nothing prevents the use of more advanced multi-kernel or ensemble strategies to further exploit the multi-kernel structure.

Let us consider the batch case first. Given $N$ training samples $(x_n,y_n)$, denote by $\vect{y}$ the vector of all $N$ outputs, and by $\vect{K}^i$ the kernel matrix corresponding to the $i$-th tap in \eqref{eq:recursive_kernel_definition}. A set of $P$ KRR models is trained as
\begin{equation}
f^i(\vect{x}) = \vect{y}^T \left( \vect{K}^i + c\vect{I} \right)^{-1} \bm{\kappa}^i(x) \,,
\end{equation}
where $\bm{\kappa}^i(x) = \left[\kappa^i(x, x_1), \ldots, \kappa^i(x, x_N) \right]$ and $c$ is a regularization constant. We combine the basic models as $h(\vect{x}) = \sum_{i=1}^P \alpha^i f^i(x)$, where the coefficients $\boldsymbol{\alpha} = [\alpha^1, \ldots, \alpha^P]^T$ are found by minimizing
\begin{equation}
\underset{\alpha^1, \ldots, \alpha^P}{\min}\left\{ \frac{1}{2}\sum_{n=1}^N \left( y_n - \sum_{i=1}^P \alpha^i f^i(x_n) \right)^2 \right\} \,,
\label{eq:stacking_optimization}
\end{equation}
which has an immediate closed form solution
\begin{equation}
\boldsymbol{\alpha} = \vect{F}^{-1}\vect{y} \,,
\end{equation}
where $\lfloor \vect{F} \rfloor_{ni} = f^i(x_n)$. This formulation is a basic form of what is known as ``stacking'' in the machine learning literature \cite{breiman1996stacked,dvzeroski2004combining}. When $P\ll N$, Eq.  \eqref{eq:stacking_optimization} does not require regularization. Otherwise, Eq. \eqref{eq:stacking_optimization} can be replaced by a more general leave-one-out strategy as in the original stacking problem \cite{breiman1996stacked}. More generally, we can include several constraints and/or regularization terms to Eq. \eqref{eq:stacking_optimization} in order to force a specific structure on $\boldsymbol{\alpha}$, such as the requirement to lie in a $P$-simplex (similar to an adaptive combination of filters \cite{arenas2016combinations}), or impose an $\ell_1$-norm regularization to remove unnecessary lags.

A similar stacking strategy can be applied in the online case. In particular, given the new datum $(x_n, y_n)$, we first update $P$ KAFs in parallel, for instance employing $P$ KLMS filters \cite{liu2008kernel}
\begin{equation}
f^i_n = f^i_{n-1} + \eta \left[ y_n - f^i_{n-1}(x_n) \right]\kappa^i(\cdot, x_n) \,,
\end{equation}
where $\eta$ is the step-size. 
Then, we update the current estimate $\boldsymbol{\alpha}_{n-1}$ of the weighting coefficients following an instantaneous descent on \eqref{eq:stacking_optimization}
\begin{equation}
\boldsymbol{\alpha}_{n} = \boldsymbol{\alpha}_{n-1} + \nu \left( y_n - \sum_{i=1}^P \alpha^i f^i_n(x_n) \right) \bm{f}_n(x_n) \,,
\label{eq:alpha_update_online}
\end{equation}
where $\nu$ is a step-size parameter and $\lfloor \bm{f}_n(\cdot) \rfloor_{i} = f_n^i(\cdot)$. 

Note that KLMS algorithms require calculating \emph{arrays} of kernel evaluations in each iteration, which is feasible by applying the recursive formula discussed in the next section. KRLS algorithms, on the other hand, require calculating entire kernel \emph{matrices}, which is less obvious in recursive settings. This, and other concepts, such as sparsity, require  more investigation.

\section{Efficient computation of the recursive kernel}
\label{sec:efficient}
We now briefly outline an implementation to compute the recursive kernel using fast operations on column vectors.

The slowest operations in Eq. \eqref{eq:recursive_kernel_definition} are the summations of the third and fourth term. In order to speed up the calculation of the last term, we define the column vector $\bm{r}^i_n$ with elements
\begin{equation}
\bm{r}^i_n(m) = \mu^2 \sum_{j=2}^{n-1} (1-\mu)^{j-1} \kappa^i(m-1, n-j).
\label{eq:efficient_row}
\end{equation}
This vector can be obtained recursively by observing that
\begin{equation}
\bm{r}^i_{n}(m) = (1-\mu) \left(\bm{r}^i_{n-1}(m) + \mu^2 \kappa^i(m-1, n-j-1)\right).
\end{equation}
A similar recursive calculation is not possible for the third term in Eq. \eqref{eq:recursive_kernel_definition}, though this term can be obtained efficiently from the elements of the convolution $\bm{\mu}_{mem} \ast \bm{k}^i_n$, where $\bm{k}^i_n(m) = \kappa^i(m, n-1)$ and $\bm{\mu}_{mem} = \mu^2 [1-\mu,(1-\mu)^2,...,(1-\mu)^N]^\top$.

\begin{figure}[t]
\centering
\includegraphics[width=.85\linewidth]{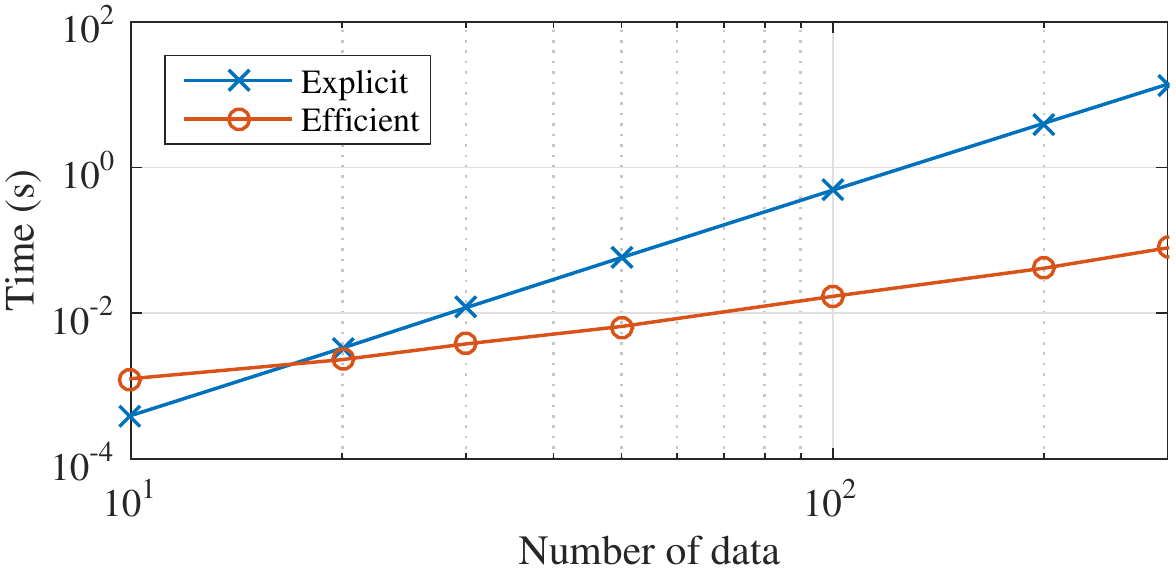}
\caption{Execution times for calculating recursive kernel matrices for different amounts of data. Explicit calculation, Eq. \eqref{eq:recursive_kernel_definition}, vs. efficient computation.}
\label{fig:speedup}
\end{figure}

Fig. \ref{fig:speedup} compares the computation times of the recursive kernel matrix with $P=5$, for an explicit implementation of Eq. \eqref{eq:recursive_kernel_definition} and the discussed efficient implementation, both in Matlab R2015a on an Intel Core i7 PC with $3.4$ GHz processor.


%
\section{Experimental results}
\label{sec:experiments}
\vspace{-1.5em}
\begin{center}
\begin{table*}
\caption{Experimental comparison of several kernels in the batch case, including a standard RBF kernel with time-delayed embeddings, the two composite recursive kernels from \cite{tuia2014explicit}, and the proposed recursive multikernel (RMK) strategies. 
}
{\centering\hfill{}
\setlength{\tabcolsep}{4pt}
\renewcommand{\arraystretch}{1.3}
\begin{small}
\begin{tabular}{lcccccc}   
\toprule
 & \textbf{Standard kernel} & \multicolumn{2}{c}{\textbf{Composite recursive kernel}} & \multicolumn{3}{c}{\textbf{Proposed recursive MK}}  \\ \cmidrule(lr){2-2} \cmidrule(lr){3-4} \cmidrule(lr){5-7}
Dataset & RBF & Average & Symmetric & SimpleMKL & Stacking & Sparse Stacking \\
\midrule
MG30 & $-18.99$ dB & $-23.26$ dB & $-22.21$ dB & \bm{$-23.42$} dB & $-20.05$ dB & $-19.92$ dB \\
Narendra & $-14.81$ dB & $-15.72$ dB & $-15.09$ dB & $-15.29$ dB & \bm{$-17.01$} dB & $-16.78$ dB \\
Wiener & \bm{$-18.20$} dB & $-17.58$ dB & $-17.23$ dB & $-17.56$ dB & $-18.19$ dB & $-18.19$ dB \\
EEG & $-5.92$ dB & $-3.69$ dB & $-3.60$ dB & $-6.23$ dB & $-6.94$ dB & \bm{$-7.69$} dB \\
Respiratory & $-18.53$ dB & $-14.92$ dB & $-12.58$ dB & $-16.82$ dB & \bm{$-18.84$} dB & $-18.64$ dB \\
EUR-USD & $-25.10$ dB & $-23.24$ dB & $-21.47$ dB & --- & \bm{$-25.83$} dB & $-25.68$ dB \\
\bottomrule
\end{tabular}
\end{small}
}
\hfill{}
\label{tab:batch_results}
\end{table*}
\end{center}

The proposed batch and online algorithms are evaluated on six benchmark problems, three of which are defined on artificially generated datasets and the rest on real-world data.


The first artificial dataset is the Mackey-Glass time-series (with delay $30$) taken from \cite{tuia2014explicit}, denoted as MG30, on which we perform one-step-ahead prediction. The second task is a noisy version of a nonlinear prediction problem introduced in \cite{narendra1990identification}, denoted as ``Narendra'', which is defined by
\begin{equation}
y_{n} = 0.3y_{n-1} + 0.6y_{n-2} + f(e_n) \,,
\end{equation}
where the unknown function $f(\cdot)$ has the form
\begin{equation}
f(e) = 0.6\sin\left( \pi e \right) + 0.3\sin\left( 3\pi e \right) + 0.1\sin\left(5\pi e\right)
\end{equation}
and $e_n = \sin\left( (1+a)\omega_0n \right)$,
%
$\omega_0 = 2\pi/250$, $a$ is uniformly distributed in the interval $\left[0.1, 2.9\right]$, and we set $y_{-1} = y_{0} = 1$. We additionally add Gaussian noise with variance $0.1$ to the desired output during training.

The third task on artificial data is the identification of the nonlinear Wiener model described in \cite{scardapane2016diffusion}, where the input is generated according to
\begin{equation}
x_n = bx_{n-1} + \sqrt{1-b^2}e_x \,,
\end{equation} 
with $x_0$ randomly generated according to a uniform distribution, $e_x$ is Gaussian noise with variance $0.1$, and we set $b=0.8$. The output is given by first applying a linear filter to an embedding of the last $8$ inputs, and then applying a soft nonlinearity on the resulting scalar value.

The first problem on real-world data considers the $4$-step ahead prediction of the EEG dataset from \cite{tuia2014explicit}, extracted from the MIT-BIH Polysomnographic Database\footnote{\url{https://www.physionet.org/physiobank/database/slpdb/}}. We then consider the prediction of a respiratory motion trace recorded at the Georgetown University Hospital\footnote{\url{http://signals.rob.uni-luebeck.de/}}, originally described in \cite{ernst2012compensating}. Finally, the EUR-USD dataset contains the EUR vs. USD exchange rates in minute intervals, taken on the days January 2nd and 5th of 2009. The task is $2$-step ahead prediction.

\subsection{Batch experiments}

In all batch experiments, we use $200$ elements for training and $1000$ separate elements for testing, except for EUR-USD, where we use $1440$ samples for training and $1370$ for testing, respectively. Parameters are fine-tuned following the same grid-search procedure described in \cite{tuia2014explicit}, optimizing over a third, independent validation set. 

We evaluate several KRR models, trained using (i) a standard RBF kernel with time-delayed embeddings on input, (ii) the recursive kernel with the two composition strategies described in \cite{tuia2014explicit} (averaging and symmetrization), and (iii) the stacking procedure described in Section \ref{sec:multikernel}, both with $\ell_2$ and $\ell_1$ regularization. Additionally, we consider an extension of our idea by training a support vector regression model with the SimpleMKL algorithm \cite{rakotomamonjy2008simplemkl}, which finds a new kernel via an adaptive combination of the base kernel matrices. Denoting by $\text{E}^2$ the mean-squared error (MSE) over the test set, we evaluate the models using a normalized MSE on a logarithmic scale,
\begin{equation}
\text{nMSE} = 10\log_{10}\left(\text{E}^2 /\hat{\sigma}_y^2\right) \,,
\end{equation}
where $\hat{\sigma}_y^2$ is the empirical variance of the output computed over the test set. 

The results for the different benchmarks are given in Table \ref{tab:batch_results}, where the best result for each dataset is highlighted with a bold font. We observe that the proposed stacking procedures are outperforming the standard KRR models in $4$ out of $6$ benchmarks, while SimpleMKL achieves slightly better performance in the Mackey-Glass task. To confirm the results, we employ the corrected Friedman test described in \cite{demvsar2006statistical}, according to which the performance of the algorithms are statistically different with a confidence value of $\alpha=0.05$. A set of Nemenyi post-hoc tests shows that the performance of the base stacking procedure is statistically better than the standard kernel and the two composite recursive kernels, while the sparse stacking procedure is statistically better than the standard kernel and the symmetric recursive kernel, only.

\subsection{Online experiments}

\begin{table}
\caption{Results for adaptive filtering with different kernels.}
{\centering\hfill{}
\setlength{\tabcolsep}{4pt}
\renewcommand{\arraystretch}{1.3}
\begin{small}
\begin{tabular}{lcc}   
\toprule
Dataset & KLMS & KLMS (RMK)\\
\midrule
MG30 & $-12.50$ dB & \bm{$-17.99$} dB \\
Narendra & $-5.65$ dB & \bm{$-14.55$} dB \\
Wiener & $-13.42$ dB & \bm{$-13.68$} dB \\
EEG & $-5.32$ dB & \bm{$-7.33$} dB \\
Respiratory & \bm{$-12.56$} dB & $-11.53$ dB \\
EUR-USD & $-21.35$ dB & \bm{$-25.13$} dB \\
\bottomrule
\end{tabular}
\end{small}
}
\hfill{}
\label{tab:online_results}
\end{table}

In a second set of experiments we evaluate the online performance of the described KLMS algorithm with recursive multikernel, and classical KLMS \cite{liu2008kernel}, on the six benchmark problems. Both algorithms use the same kernel parameter and learning rate. Table \ref{tab:online_results} lists the nMSE results obtained after convergence, indicating clear benefits of the RMK strategy.

\begin{figure}[t]
\centering
\includegraphics[width=.95\linewidth]{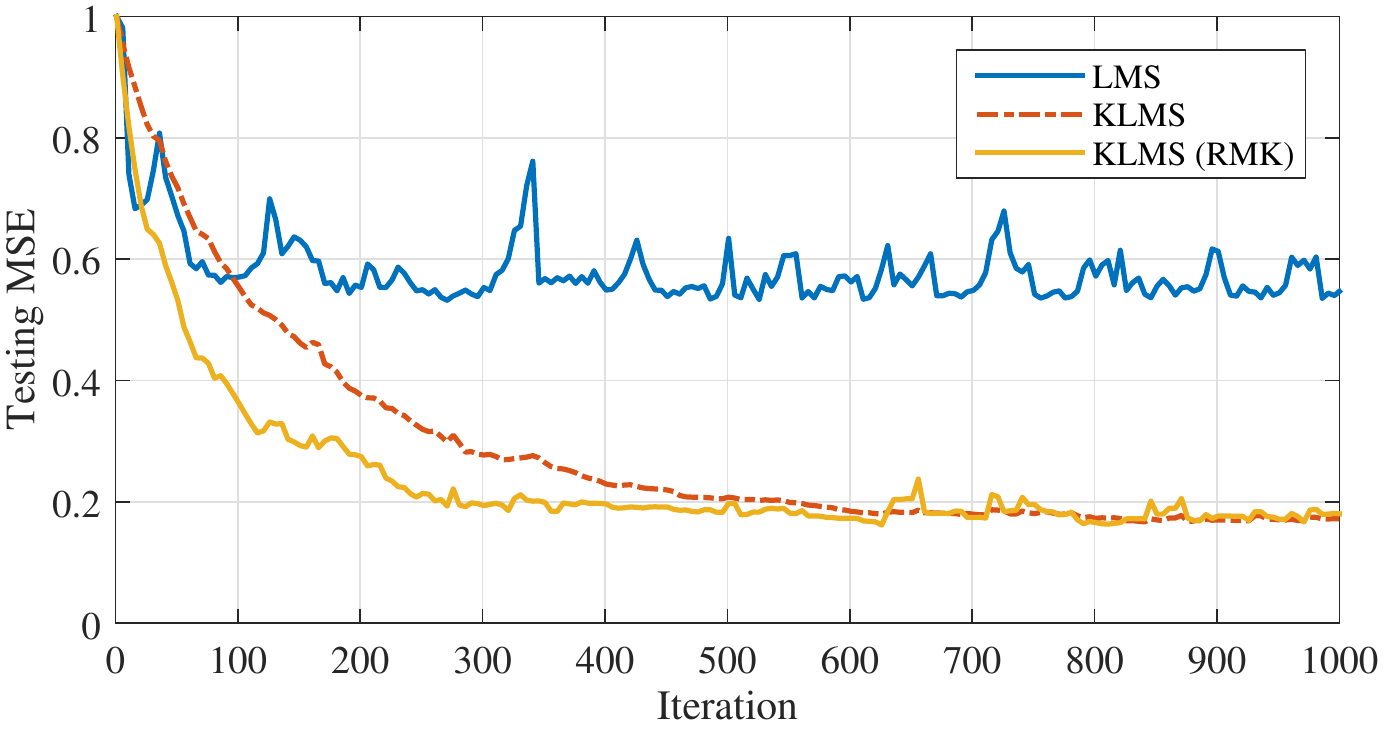}
\caption{Learning curves for the channel equalization experiment.}
\label{fig:convergence}
\end{figure}

Finally, in Fig. \ref{fig:convergence} we reproduce the online learning experiment from \cite[Sec. 2.11.2]{liu2008kernel}. In this experiment, a binary signal is fed into a nonlinear communications channel, and the goal is to estimate the correct input signal given its output, i.e. to construct a \emph{channel equalizer}. KLMS with RMK employs $P=5$ lags and $\mu=0.9$, and the same kernel parameter and learning rate as standard KLMS. While both nonlinear algorithms converge to a similar MSE, the RMK algorithm enjoys a much faster convergence rate.

\section{Conclusions}
\label{sec:conclusions}
In this paper, we proposed a novel approach for exploiting multiple time-scale information in kernel regression and filtering problems by combining a previously introduced $\gamma$-filter (defined in a proper Hilbert space), with an adaptive strategy for combining kernel functions. In this way, the algorithm automatically adapts to the most significant time-scales of the original input signal. Additionally, the kernel functions are particularly suitable for online processing because they can be recursively computed from previous values, and we briefly discussed on their efficient implementation. 

Experimental simulations show that the algorithm has similar or better performance on a wide range of tasks, when compared to several alternative strategies such as time-delayed embeddings of the input. In future works, we plan to further extend our idea by leveraging upon the recent literature on MK filters, and by exploring nonlinear combinations of the different kernels.

\section*{Acknowledgment}
S. Van Vaerenbergh is supported by the Spanish Ministry of Economy and Competitiveness (under project TEC2014-57402-JIN). S. Scardapane is supported in part by Italian MIUR, ``\textit{Progetti di Ricerca di Rilevante Interesse Nazionale}'',  GAUChO project, under Grant 2015YPXH4W\_004.



\bibliographystyle{IEEEtran}
\bibliography{biblio}

\end{document}